\documentclass[sigconf]{acmart}

\usepackage{hyperref}
\usepackage{url}
\usepackage{booktabs}       % professional-quality tables
\usepackage{amsfonts}       % blackboard math symbols
\usepackage{nicefrac}       % compact symbols for 1/2, etc.
\usepackage{microtype}      % microtypography
\usepackage{xcolor}         % colors
\usepackage{natbib}
\usepackage{multirow}
\usepackage{amsmath}
\usepackage{graphicx}
\usepackage{colortbl}
\usepackage{arydshln}
\usepackage{wrapfig}
\usepackage{enumitem}
\usepackage{subcaption}
\usepackage{balance}

\usepackage{xspace}
\newcommand{\SDE}{\texttt{SDE}\xspace}

\AtBeginDocument{%
  }

\setcopyright{acmlicensed}
\copyrightyear{2018}
\acmYear{2018}
\acmDOI{XXXXXXX.XXXXXXX}
\acmConference[Conference acronym 'XX]{Make sure to enter the correct
  conference title from your rights confirmation email}{June 03--05,
  2018}{Woodstock, NY}
\acmISBN{978-1-4503-XXXX-X/2018/06}

\copyrightyear{2026}
\acmYear{2026}
\setcopyright{cc}
\setcctype{by}
\acmConference[WWW '26]{Proceedings of the ACM Web Conference 2026}{April 13--17, 2026}{Dubai, United Arab Emirates}
\acmBooktitle{Proceedings of the ACM Web Conference 2026 (WWW '26), April 13--17, 2026, Dubai, United Arab Emirates}
\acmPrice{}
\acmDOI{10.1145/3774904.3792528}
\acmISBN{979-8-4007-2307-0/2026/04}

\begin{document}

\title{Spectral Disentanglement and Enhancement: A Dual-domain Contrastive Framework for Representation Learning}

\author{Jinjin Guo}
\affiliation{
  \institution{JD.com}
  \city{Beijing}
  \country{China}}
\email{umguojinjin@outlook.com}

\author{Yexin Li }
\authornote{Corresponding author}
\affiliation{
  \institution{State Key Laboratory of General Artificial Intelligence, BIGAI}
  \city{Beijing}
  \country{China}}
\email{liyexin@bigai.ai}

\author{Zhichao Huang}
\affiliation{
  \institution{JD.com}
  \city{Beijing}
  \country{China}}
\email{iceshzc@gmail.com}

\author{Jun Fang}
\affiliation{
  \institution{JD.com}
  \city{Beijing}
  \country{China}}
\email{fangjun8@jd.com}

\author{Zhiyuan Liu}
\affiliation{
  \institution{JD.com}
  \city{Beijing}
  \country{China}}
\email{liuzhiyuan8@jd.com}
  
\author{Chao Liu}
\affiliation{
  \institution{JD.com}
  \city{Beijing}
  \country{China}}
\email{liuchao397@jd.com}

\author{Pengzhang Liu}
\affiliation{%
  \institution{JD.com}
  \city{Beijing}
  \country{China}}
\email{liupengzhang@jd.com}

\author{Qixia Jiang}
\affiliation{%
  \institution{JD.com}
  \city{Beijing}
  \country{China}}
\email{jiangqixia@jd.com}

\renewcommand{\shortauthors}{Jinjin Guo et al.}
%% No italics, no superscripts, not anonymous
%% Use footnote or author note to identify equal contribution and/or contact author info

\begin{abstract}
Large-scale multimodal contrastive learning has recently achieved impressive success in learning rich and transferable representations, yet it remains fundamentally limited by the uniform treatment of feature dimensions and the neglect of the intrinsic spectral structure of the learned features. Empirical evidence indicates that high-dimensional embeddings tend to collapse into narrow cones, concentrating task-relevant semantics in a small subspace, while the majority of dimensions remain occupied by noise and spurious correlations. Such spectral imbalance and entanglement undermine model generalization. We propose Spectral Disentanglement and Enhancement (\SDE), a novel framework that bridges the gap between the geometry of the embedded spaces and their spectral properties. Our approach leverages singular value decomposition to adaptively partition feature dimensions into strong signals that capture task-critical semantics, weak signals that reflect ancillary correlations, and noise representing irrelevant perturbations. A curriculum-based spectral enhancement strategy is then applied, selectively amplifying informative components with theoretical guarantees on training stability. Building upon the enhanced features, we further introduce a dual-domain contrastive loss that jointly optimizes alignment in both the feature and spectral spaces, effectively integrating spectral regularization into the training process and encouraging richer, more robust representations. Extensive experiments on large-scale multimodal benchmarks demonstrate that \SDE consistently improves representation robustness and generalization, outperforming state-of-the-art methods. \SDE integrates seamlessly with existing contrastive pipelines, offering an effective solution for multimodal representation learning.
\end{abstract}




\maketitle

\section{Introduction}
\label{sec_intro}
Learning robust and generalizable representations is a fundamental challenge in deep learning, especially for large-scale neural networks trained with massive data.

Contrastive learning \cite{gutmann2010noise,hadsell2006dimensionality} has emerged as a powerful paradigm for learning representations in both unsupervised and supervised settings, achieving state-of-the-art performance across computer vision \cite{chen2020simple,10.5555/3495724.3497510,caron2020unsupervised,chen2021exploring}, natural language understanding \cite{gao2021simcse,li2023gte}, and multimodal tasks \cite{radford2021clip,jia2021align,zhai2023sigmoid,li2022blip,wei2024uniir,ren2024vista,zhang2024magiclens}. Recent advances have further extended contrastive learning frameworks by leveraging synthetic data generation with large language models (LLMs)~\cite{chen2025mme5,zhang2024gme,zhou2024megapairs}, as well as feature-level interventions, such as adversarial noise injection, e.g., CLAE~\cite{ho2020contrastive}, SimCSE~\cite{gao2021simcse}, and NEFTune~\cite{jain2023neftune}, to improve robustness.

Despite these advances, a critical limitation remains: most existing methods treat all feature dimensions uniformly. Empirical studies have shown that effective embedding spaces are often confined within narrow cones~\cite{liang2022mind}. For instance, in a 512-dimensional embedding space, vectors with a cosine similarity of 0.56—when projected onto the 512-dimensional unit hypersphere-are confined to a very tiny surface area, indicating that this seemingly moderate similarity originates from a highly concentrated region of the original feature manifold. This suggests that the effective dimensionality of learned representations is much lower than the nominal one \cite{liang2022mind}.

% Recent work~\cite{nayak2025sculpting,hartford2024spectrum} further supports this observation, showing that neural network parameters exhibit substantial redundancy, where only a small subset of spectral directions, i.e., those associated with large singular values, encode critical task-specific knowledge, while the remaining dimensions contribute minimally. This spectral decomposition aligns with the narrow cone phenomenon in embedding space, suggesting that both parameter and representation spaces are intrinsically low-dimensional, dominated by a few salient directions. 

% Our work bridges the gap between empirical observations of narrow embedding cones and the spectral properties of neural representations. While prior studies \cite{nayak2025sculpting,hartford2024spectrum} identified parameter-space redundancy via SVD, we reveal three key limitations of current multimodal contrastive learning approaches:

Given this observation, we identify several key limitations of current multimodal representation learning approaches:

\begin{itemize}[leftmargin=2em]
\item \textbf{Spectral imbalance}. Dominant semantic features concentrate in a small subspace, while weaker signals and noise occupy the majority of dimensions, creating anisotropic embeddings that can complicate downstream optimization.

\item \textbf{Uniform optimization pitfalls}. Standard multimodal contrastive learning approaches treat feature dimensions equally, leading to the following issues:
\begin{itemize}
    \item Entangled semantics—task-relevant features become inseparable from ancillary correlations and noise;
    \item Suboptimal robustness—noise perturbations propagate indiscriminately across all feature dimensions;
    \item Limited generalization—weak or spurious correlations are amplified during optimization.
\end{itemize}
\end{itemize}

While prior studies~\cite{nayak2025sculpting,hartford2024spectrum} leverage singular value decomposition (SVD) to identify redundancy in neural networks, they focus exclusively on the parameter space, leaving the feature space largely unexplored. Our work bridges this gap by connecting empirical observations of narrow embedding cones with the spectral properties of neural representations.

%Although prior work has employed Singular Value Decomposition (SVD) to investigate parameter spaces~\cite{nayak2025sculpting,hartford2024spectrum}, the spectral structure of feature spaces in contrastive learning remains largely unexamined. In practice, feature embeddings naturally decompose into distinct spectral components, i.e., strong signals, which capture dominant task-critical semantics; weak signals, corresponding to ancillary correlations; and noise, representing task-irrelevant perturbations. Neglecting this intrinsic structure can lead to biased optimization and impaired generalization.

%这里需要插入图片
%\begin{figure}[t]
%   \centering
%    \includegraphics[width=0.95\textwidth]{ASP_framework}
%    \caption{Curriculum Spectral Partition (ASP) Framework: (a) Standard contrastive learning treats all feature dimensions uniformly; 
%   (b) ASP dynamically decomposes features via SVD into strong/weak/noise subspaces; 
%    (c) Curriculum strategy progressively enhances signal subspaces while suppressing noise; 
%    (d) Dual-domain loss operates in both feature and spectral spaces.}
%    \label{fig:framework}
%\end{figure}

We propose the \SDE framework, where feature embeddings are dynamically disentangled and enhanced according to their spectral properties. \SDE introduces three key innovations:
(1) Spectral Disentanglement — real-time SVD is employed to partition features into strong, weak, and noise subspaces during multimodal contrastive training;
(2) Spectral Enhancement — a refinement strategy that selectively amplifies dominant signals, normalizes weaker ones, and suppresses noise;
(3) Dual-Domain Contrastive Learning — jointly regularizes alignment in both the feature and spectral spaces, thereby enhancing the robustness and generalization of the learned representations.
\SDE is the first study to introduce spectral-aware feature alignment within a contrastive learning framework, establishing a practical and versatile foundation for robust multimodal representation learning.

Extensive experiments demonstrate that \SDE achieves superior performance on the multimodal benchmark MMEB~\cite{jiang2024vlm2vec} compared with state-of-the-art baselines, validating its effectiveness when integrated into existing multimodal contrastive learning pipelines.

In summary, our contributions are fourfold:
\begin{itemize}[leftmargin=2em]
    \item We identify the spectral structure of feature representations as a critical bottleneck in multimodal representation learning, showing that uniform optimization across feature dimensions can obscure task-relevant semantics with noise.
    
    \item We propose an adaptive spectral disentanglement and enhancement method, in which feature embeddings are decomposed via SVD and adaptively refined through curriculum-based spectral manipulation.
    
    \item We design a dual-domain contrastive loss that simultaneously enforces alignment across both the feature and spectral spaces, yielding more robust and generalizable representations.
    
    \item Extensive experiments on MMEB demonstrate that our approach consistently outperforms state-of-the-art baselines, demonstrating the effectiveness of spectral-aware feature disentanglement and enhancement.
\end{itemize}

\section{Related Work}
\label{sec_related}

\paragraph{Contrastive Representation Learning}
Contrastive representation learning (CRL) aims to learn an embedding space in which similar instances are closer than dissimilar ones, and has achieved remarkable success across vision~\cite{chen2020simple, 10.5555/3495724.3497510, caron2020unsupervised, chen2021exploring}, language~\cite{reimers2019sentence, gao2021simcse}, and graph domains~\cite{you2020graph, qiu2020gcc}. A key factor underlying the success of CRL—particularly in vision—lies in carefully designed data augmentation strategies that generate diverse yet semantically consistent positive views, such as random cropping~\cite{chen2020simple} and node or edge dropping or diffusion~\cite{you2020graph, zhu2020graph}.
Conversely, effective negative sampling, especially the identification of hard negatives that are semantically similar to but distinct from the anchor, is crucial for learning discriminative representations. Negative mining strategies have evolved from random sampling to hard negative mining via similarity ranking~\cite{robinson2020contrastive}, adversarial generation~\cite{Kalantidis2020HardNM}, and memory bank-based approaches~\cite{he2020momentum}. InfoNCE~\cite{oord2018representation} remains the predominant contrastive objective, maximizing mutual information through temperature-scaled similarity. Notable extensions include debiased objectives that mitigate false negatives~\cite{chuang2020debiased} and feature decorrelation losses that eliminate the need for explicit negative samples~\cite{zbontar2021barlow, bardes2021vicreg}.

\paragraph{Multimodal Representation Learning} 
Multimodal models learns to map heterogeneous modalities into a shared representation space~\cite{pmlr-v182-zhang22a, jia2021scaling}. Early approaches typically adopt dual-encoder architectures to align modality-specific features via contrastive objectives, exemplified by CLIP~\cite{radford2021clip} and ALIGN~\cite{jia2021align}. More recent work explores deeper integration of visual–linguistic representations: UniIR~\cite{wei2024uniir} combines independently encoded embeddings through score fusion, while VISTA~\cite{ren2024vista} augments text encoders with visual fusion modules.
Despite achieving strong cross-modal retrieval performance, these methods often struggle to capture fine-grained semantic alignment, which limits their effectiveness on more complex multimodal tasks~\cite{zhang2024gme}. Vision–Language Models (VLMs) have recently emerged as dominant backbones for multimodal embeddings, owing to their transformer-based architectures that intrinsically fuse cross-modal features. Models such as Qwen-VL~\cite{wang2024qwen2}, Phi~\cite{abdin2024phi}, and LLaVA-Next~\cite{liu2024llavanext} demonstrate strong generalization across diverse image–text compositions, particularly in instruction-following settings. This progress has spurred a growing line of VLM-based embedding methods. For example, E5-V~\cite{jiang2024e5v} and VLM2Vec~\cite{jiang2024vlm2vec} convert VLMs into embedding models via instruction tuning. LLaVE~\cite{lan2025llave} introduces hardness-weighted contrast with reward models to prioritize difficult negatives, while UniME~\cite{gu2025breaking} employs a two-stage pipeline with textual knowledge distillation and hard-negative tuning.
In parallel, universal representation learning seeks modality-agnostic embeddings: MAGICLens~\cite{zhang2024magiclens} constructs multimodal knowledge graphs for alignment, whereas Universal~\cite{liu2022universal} develops cross-modal attention mechanisms.

\paragraph{Spectral Analysis} Building on recent advances in multimodal representation learning, spectral analysis has emerged as a powerful tool for uncovering the fundamental properties of learned features. Empirical studies have revealed that the feature spaces of large-scale multimodal models often exhibit a pronounced cone effect, where representations are confined to narrow regions of the embedding space~\cite{liang2022mind}. Further investigations~\cite{hartford2024spectrum,sharma2023truth} demonstrate that neural network parameters possess substantial spectral redundancy: directions associated with small singular values contribute little to critical model knowledge, whereas dominant singular directions encode task-relevant semantics.
This spectral perspective offers new insights into the internal structure of learned representations. Specifically, feature embeddings can be naturally decomposed into distinct spectral bands, corresponding to strong semantic signals, ancillary correlations, and stochastic noise. Such decomposition enables principled strategies for disentangling useful information from irrelevant or noisy components. Recent work has leveraged this perspective in parameter space. For example, the locate-and-edit paradigm first identifies influential parameters and then selectively perturbs them to improve robustness or adaptivity. The recent study ~\cite{nayak2025sculpting} further demonstrates that parameter spaces can be partitioned into task-specific and non-task-specific subspaces, facilitating continual learning.
However, these advances focus exclusively on parameter space, leaving an open gap in feature space analysis. Our work bridges this gap by introducing the first feature-space spectral disentanglement framework, combining insights from parameter analysis with novel adaptations for multimodal representation learning.

\section{Methodology}
\label{sec_method}

\begin{figure*}[ht]
    \centering
    \includegraphics[width=0.9\linewidth, trim={2.0cm 3.0cm 2.0cm 1.4cm}, clip]{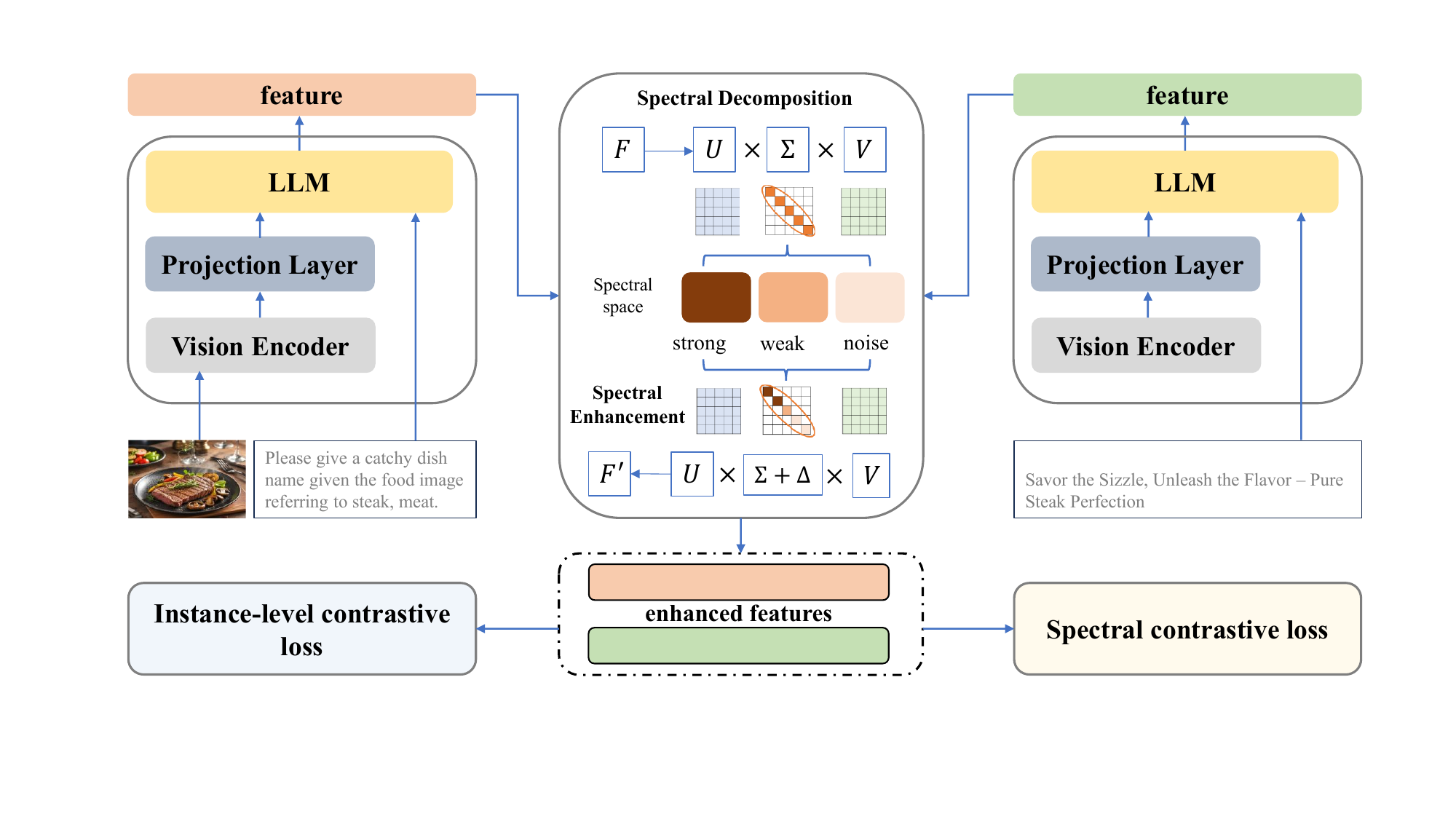}
    \caption{Overview of the \SDE framework. The VLM backbone jointly encodes the query, consisting of an image and text, and the target text input to produce multimodal feature representations. These features are then partitioned into strong, weak, and noise subspaces via SVD, with each subspace adaptively enhanced and reconstructed back to the feature space. Finally, a dual-domain contrastive loss—comprising instance-level alignment in the feature space and structure-aware alignment in the spectral space—is applied to improve both robustness and generalization.}
    \label{fig:main_frame}
\end{figure*}

In this section, we give a detailed description of \SDE as illustrated by Fig.~\ref{fig:main_frame}. It is a novel framework designed to disentangle and enhance the spectral structure of the feature space within multimodal contrastive learning. It consists of three core components: adaptive spectral disentanglement, curriculum-based spectral enhancement, and dual-domain contrastive loss, which are given in the following.

\subsection{Spectral Disentanglement}
Suppose $\mathbf{F} \in \mathbb{R}^{m \times n}$ is the feature matrix with $m$ samples and $n$ feature dimensions, obtained from the backbone of a vision language model (VLM). We apply SVD to analyze its spectral properties, yielding:
\begin{equation}
\label{eq:svd}
    \mathbf{F} = \mathbf{U} \mathbf{\Sigma} \mathbf{V}^\top;
\end{equation}
where $\mathbf{U} \in \mathbb{R}^{m \times m}$ and $\mathbf{V} \in \mathbb{R}^{n \times n}$ are unitary matrices, and $\mathbf{\Sigma} \in \mathbb{R}^{m \times n}$ contains non-negative singular values $\sigma_1 \geq \sigma_2 \geq \cdots \geq \sigma_r > 0$, with $r = \text{rank}(\mathbf{F})$. Columns of $\mathbf{V}$ define basic directions of the features, while singular values quantify directional energy concentrations.

Empirical studies \cite{hartford2024spectrum, sharma2023truth} have shown that a substantial proportion of the feature space exhibits redundancy. Specifically, those associated with smaller singular values often reflect redundancy or noise. Conversely, directions corresponding to larger singular values tend to encode the most salient and task-relevant information.

Inspired by this observation, we propose to disentangle the feature space, partitioning it into three distinct subspaces according to the magnitude of the singular values:

\begin{itemize}[leftmargin=2em]
\item \textbf{Strong signal}: dominant and task-specific feature dimensions that are crucial for model performance.
\item \textbf{Weak signal}: subtle but potentially informative variations that may contribute to understanding or generalization.
\item \textbf{Noise}: random fluctuations or irrelevant variations, often detrimental to robustness.
\end{itemize}

We determine the thresholds that define the strong, weak, and noise subspaces using the interquartile range (IQR) of the singular values~\cite{hartford2024spectrum}. Specifically, we employ the Marchenko–Pastur distribution~\cite{marvcenko1967distribution,hartford2024spectrum} to establish principled bounds on the eigenvalues of random covariance matrices, thereby guiding the partitioning of the feature space.

For a matrix $\mathbf{F}$ whose entries are independent and identically distributed with zero mean and standard deviation $\vartheta$, the eigenvalues of $ C = \frac{1}{n} \mathbf{F}^\top \mathbf{F} $ converge to a distribution supported on $[\vartheta^{2}(1 - \sqrt{m/n})^2$, $\vartheta^{2}(1 + \sqrt{m/n})^2]$. Accordingly, the singular values of $\mathbf{F}$ asymptotically satisfy:

\begin{equation}
\sigma_i \in \left[ \vartheta (\sqrt{n} - \sqrt{m}), \vartheta (\sqrt{n} + \sqrt{m}) \right];
\end{equation}
where $i =1, 2, 3, \ldots, r$. These theoretical bounds facilitate the estimation of singular value quantiles, enabling partitioning of the feature dimensions into the aforementioned subspaces.

Spectral disentanglement enables the separation of informative features from less useful or noisy ones during training. By explicitly identifying and partitioning these subspaces, our \SDE framework lays the groundwork for enhanced representation learning.

\subsection{Spectral Enhancement}                                       
Building upon the disentanglement of the feature space, we propose a tailored spectral enhancement strategy that adaptively manipulates singular values within each subspace to enhance model robustness and generalization. For the feature matrix $\mathbf{F} \in \mathbb{R}^{m \times n}$ and its SVD $\mathbf{F} = \mathbf{U} \Sigma \mathbf{V}^\top$, where $\mathbf{\Sigma}=\text{diag}(\sigma_1, \sigma_2, \ldots, \sigma_r)$, the \textbf{enhanced feature matrix} is reconstructed as:

\begin{equation}
    \mathbf{F'} = \mathbf{U} (\mathbf{\Sigma} + \mathbf{\Delta}) \mathbf{V}^\top;
\end{equation}
where $\mathbf{\Delta}=\text{diag}(\delta_1, \delta_2,...,\delta_r)$ is a diagonal perturbation matrix designed with distinct subspace-specific strategies. 

\paragraph{Strong Signal Enhancement} Singular values in the strong signal subspace represent principal directions containing dominant, task-relevant information. To improve robustness and mitigate overfitting in these directions, we inject controlled adversarial noise into their magnitudes. The scale of this noise is adaptively determined by both the training progress and the relative strength of each singular value:

\begin{align}
\delta_i &= \alpha(t) \cdot \gamma_{\text{strong}} \cdot \epsilon_i, \quad \epsilon_i \sim N(0,1), \quad \text{if } i \in \text{strong signal}.
\end{align}
Here, $N(0,1)$ denotes a standard normal distribution with mean $0$ and variance $1$; the hyperparameter is defined as $\gamma_{\text{strong}} = \frac{\sigma_i}{\sigma_1} $, where $\sigma_1$ is the largest singular value, providing a natural normalization. The curriculum factor $\alpha(t) \in (0, 1)$ evolves with the training step $t$, implementing a curriculum learning strategy in which strong augmentation is applied in the early stages, moderate augmentation in the middle stages, and weak augmentation toward the end of training. Further details are provided in the Appendix.

\paragraph{Weak Signal Normalization} The weak signal subspace comprises feature dimensions associated with subtle yet potentially informative variations. To retain valuable diversity while maintaining stability, we employ spectral normalization, in which the weak singular values are adaptively attenuated according to their relative magnitudes and the current curriculum factor. This selective normalization encourages the model to leverage subtle but potentially informative features without amplifying noise or irrelevant fluctuations, as defined below:

\begin{align}
\delta_i &= -\alpha(t) \cdot \gamma_{\text{weak}} \cdot \sigma_i, \quad \text{if } i \in \text{weak signal};
\end{align}
where $\gamma_{\text{weak}}=\left(\frac{\sigma_i}{\sigma_1}\right)^2$ controls the normalization intensity. This choice suppresses weak singular values more aggressively. 
% preserving strong signals and encouraging the model to leverage fine-grained features without amplifying noise or irrelevant fluctuations.

\paragraph{Noise Suppression} Singular values identified as noise typically correspond to directions occupied by corrupting influences or irrelevant variations. To suppress these values, especially during the initial training phase, we adopt signal-to-noise regularization  \cite{hartford2024spectrum}. The suppression degree is modulated by the estimated signal-to-noise ratio and the training progress, allowing the regularization to relax as learning stabilizes:

\begin{align}
\delta_i &= -\alpha(t) \cdot \gamma_{\text{noise}} \cdot \sigma_i, \quad \text{if } i \in \text{noise}.
\end{align}
Here, 
$\gamma_{\text{noise}} = \frac{\sum_{i \in \mathcal{S} \cup \mathcal{W}} \delta_i}{\sum_{k \in \mathcal{N}} \delta_{k}}$, 
where $\mathcal{S}$, $\mathcal{W}$, and $\mathcal{N}$ denote the strong, weak, and noise subspaces, respectively. 
It quantifies the normalized energy ratio between the signal subspaces $\mathcal{S} \cup \mathcal{W}$ and the noise subspace $\mathcal{N}$. 
A larger $\gamma_{\text{noise}}$ indicates a higher proportion of signal energy relative to the noise, enabling more aggressive noise suppression.

%Formally, the complete spectral enhancement can be summarized as:
%\begin{equation}
%\sigma_i \leftarrow \text{clip}\left(\sigma’i, \epsilon{\text{min}}, 1.2\sigma_1\right)
%\end{equation}
\paragraph{Perturbation Analysis}
In summary, the overall enhancement can be expressed as $\boldsymbol{\Sigma'} = \boldsymbol{\Sigma} + \boldsymbol{\Delta}$, where the diagonal perturbation matrix is defined as follows:

\begin{equation}
\label{eq:delta}
\boldsymbol{\Delta}_{i} = \alpha(t)
\begin{cases}
 \frac{\sigma_i}{\sigma_1} \cdot \epsilon_i  & \text{if } i \in \text{strong} \\
- \left(\frac{\sigma_i}{\sigma_1}\right)^2 \sigma_i & \text{if } i \in \text{weak} \\
- \gamma_{\text{noise}} \cdot \sigma_i & \text{if } i \in \text{noise}.
\end{cases}
\end{equation}

The expected Frobenius norm of the perturbation matrix $||\boldsymbol{\Delta}||_{F}^{2} = \sum_{i=1}^r |\delta_i|^2$ satisfies the bound:

\begin{align}
\mathbb{E} \left[ ||\boldsymbol{\Delta}||_{F}^{2} \right] \leq \alpha^2(t) \left[ |\mathcal{S}| + \sum_{j \in \mathcal{W}} \left(\frac{\sigma_j}{\sigma_1}\right)^4 \sigma_j^2 + \gamma^2_{\text{noise}} \sum_{k \in \mathcal{N}} \sigma_k^2 \right];
\end{align}
where $|\mathcal{S}|$ is the number of strong signals, and the adversarial noise $\epsilon$ is assumed to follow a standard Gaussian distribution, i.e., $\mathbb{E}[\epsilon^2]=1$. 

In practice, since the singular values $\sigma_i$ monotonically decrease, the cumulative energy in both the weak and noise subspaces remains quite limited. 
As a result, the spectral enhancement induces well-controlled perturbations governed by the curriculum factor $\alpha(t)$. This ensures the spectral enhancement introduces minimal instability, while adaptively promoting discriminative feature learning, suppressing noise, and improving robustness and generalization throughout the training process.

\subsection{Dual-domain Contrastive Learning}
In this subsection, to fully exploit the benefits of spectral disentanglement and enhancement, we propose a \textbf{dual-domain optimization paradigm} that simultaneously aligns representations in both the \textbf{feature} space and the \textbf{spectral} space. This approach not only encourages instance-level alignment but also enforces global consistency of the spectral structure, leading to more robust and semantically meaningful representations.

\paragraph{Instance-level Alignment} Traditional contrastive learning methods focus on instance-level alignment in the feature space, where similar samples are pulled closer while the dissimilar ones are pushed apart. The InfoNCE loss \cite{oord2018infonce} is widely adopted to achieve this, operating on a query vector $\mathbf{X} \in \mathcal{R}^{m\times n}$ and its corresponding target vector $\mathbf{Y}\in \mathcal{R}^{m\times n}$, which is denoted as:

\begin{equation}
\label{eq:infonce}
   \mathcal{L}_{\text{feat}}(\mathbf{X},\mathbf{Y}) = -\sum_{i=1}^m \log \frac{\exp(s(\mathbf{x}_i, \mathbf{y}_i) / \tau)}{\sum_{j=1}^m \exp(s(\mathbf{x}_i, \mathbf{y}_j) / \tau)};
\end{equation}
where $s(\mathbf{a}, \mathbf{b}) = \frac{\mathbf{a}^\top \mathbf{b}}{{\left\| \mathbf{a} \right\|}_{2}  {\left\| \mathbf{b} \right\|}_{2}}$ is the cosine similarity,  $\tau$ is the temperature hyperparameter, and $m$ is the batch size.

However, instance-level alignment alone is insufficient for capturing global structural consistency. Specifically, it is vulnerable to adversarial manipulations via orthogonal transformations, which preserve pairwise distances but can distort the semantic structure. For any orthogonal matrix $\mathbf{Q}$ with $\mathbf{Q} \mathbf{Q}^\top = \mathbf{I}$, we have

\begin{equation}
\label{eq:orth_trans}
s(\mathbf{Qx}, \mathbf{Qy}) = \frac{(\mathbf{Qx})^\top (\mathbf{Qy})}{{\left\| \mathbf{Qx} \right\|}_{2} {\left\| \mathbf{Qy} \right\|}_{2}} 
= \frac{\mathbf{x}^\top\mathbf{Q}^\top\mathbf{Qy}}{{\left\| \mathbf{x} \right\|}_{2} {\left\| \mathbf{y} \right\|}_{2}} 
= s(\mathbf{x}, \mathbf{y}).
\end{equation}

This property implies that adversarial samples generated via orthogonal transformations can distort semantic structures while fooling instance-level objectives, as they preserve local distances but disrupt global geometry. Consequently, solely relying on instance-level alignment may fail to capture high-level structural consistency between modalities.

\paragraph{Spectral Contrastive Loss} To address the above limitation, we introduce a spectral contrastive loss that explicitly enforces alignment from a spectral perspective. This loss comprises two complementary components:

\begin{enumerate}
\item \textbf{Spectral Distribution Alignment:}
The singular value vector $\boldsymbol{\sigma}$ characterizes the energy distribution across latent dimensions. Aligning $\boldsymbol{\sigma_x}$ and $\boldsymbol{\sigma_y}$ ensures both modalities exhibit similar feature importance across dimensions. The normalized spectral distribution is defined as:

\begin{align}
\begin{split}
\mathbf{p}_X &= \frac{\mathbf{w} \odot \boldsymbol{\sigma}_X}{\|\mathbf{w} \odot \boldsymbol{\sigma}_X\|_2}, \quad
\mathbf{p}_Y = \frac{\mathbf{w} \odot \boldsymbol{\sigma}_Y}{\|\mathbf{w} \odot \boldsymbol{\sigma}_Y\|_2};
\end{split} 
\end{align}
where $\mathbf{w}$ is a predefined linearly decaying weighting vector that prioritizes the alignment of leading singular values, and $\odot$ denotes the element-wise product. The \textit{Hellinger distance} then quantifies the discrepancy between $\mathbf{p}_X$ and $\mathbf{p}_Y$:

\begin{equation}
   \mathcal{L}_{\text{hellinger}} = \frac{1}{\sqrt{2}} \left\| \sqrt{\mathbf{p}_X} - \sqrt{\mathbf{p}_Y} \right\|_2 .
\end{equation}

\item \textbf{Subspace Consistency Enforcement:}  
While spectral alignment controls dimension-wise importance, subspace alignment further ensures directional coherence of principal components. The top-$k$ singular vectors $\mathbf{V}$ span the dominant semantic subspace. Minimizing their Gram matrix deviation enforces orthonormal alignment:

\begin{align}
\mathbf{G} &= \mathbf{V}_X^\top \mathbf{V}_Y  \\
\mathcal{L}_{\text{subspace}} &= \frac{1}{\sqrt{2k}} \left\| \mathbf{G} - \mathbf{I}_k \right\|_F;
\end{align}
where $\mathbf{I}_k$ is $k$-dimensional identity matrix.
This ensures that the principal directions remain consistently aligned across modalities. Without such a constraint, orthogonal transformations could arbitrarily rotate these subspaces while preserving instance-level distances, thereby undermining semantic coherence.
\end{enumerate}

As shown, the necessity of spectral alignment and subspace consistency arises from the fact that orthogonal transformations preserve pairwise distances but can arbitrarily alter both the spectral energy distribution and principal directions. By aligning both singular value distributions and principal subspaces, our framework enforces a more holistic and robust alignment. The composite spectral loss is then defined as:

\begin{equation}
\mathcal{L}_{\text{spec}} = \frac{1}{2} \left( \mathcal{L}_{\text{hellinger}} + \mathcal{L}_{\text{subspace}} \right).
\end{equation}

Consequently, our dual-domain objective integrates both the instance-level and spectral structure alignment:

\begin{equation}
\label{eq:total_loss}
\begin{split}
\mathcal{L}_{\text{total}} &= \underbrace{\mathcal{L}_{\text{feat}}(\mathbf{X}_{\text{enhanced}}, \mathbf{Y}_{\text{enhanced}})}_{\text{instance alignment}} +   \\
& \lambda(t)\cdot \underbrace{\mathcal{L}_{\text{spec}}(\mathbf{X}_{\text{enhanced}}, \mathbf{Y}_{\text{enhanced}})}_{\text{spectral structure alignment}};
\end{split}
\end{equation}
where $\lambda(t)$ is a dynamically scheduled weighting coefficient that adapts throughout training. $\lambda(t)$ gradually decreases, shifting emphasis toward fine-grained instance-level alignment—details of which are provided in the Appendix. This adaptive scheduling ensures robust convergence and prevents over-regularization. Here, $\mathbf{X}_{\text{enhanced}}$ and $\mathbf{Y}_{\text{enhanced}}$ denote the features obtained after spectral disentanglement and enhancement, corresponding to $\mathbf{X}$ and $\mathbf{Y}$, respectively. 

By jointly optimizing for instance-level discrimination and global spectral consistency, our dual-domain contrastive learning framework leverages the strengths of both feature and structure alignment. This design not only mitigates the vulnerability of instance-level objectives to adversarial orthogonal transformations but also ensures that the learned representations capture both local and global geometric properties for robustness and generalization.

\section{Experiment}
\label{sec_exp}
In this section, we comprehensively evaluate our proposed \SDE framework on the well-established Massive Multimodal Embedding Benchmark (MMEB)~\cite{jiang2024vlm2vec}. We compare \SDE with recent state-of-the-art baselines, evaluating its effectiveness and generalization across a wide range of multimodal tasks, including classification, visual question answering (VQA), retrieval, and visual grounding.

Due to space limitations, the main text focuses on the core experimental settings and results, while additional implementation details and results are provided in the Appendix.

\subsection{Experimental Setup}
\paragraph{Dataset Overview} MMEB~\cite{jiang2024vlm2vec} is a large-scale, task-diverse benchmark designed for evaluating multimodal embedding models. It comprises 36 datasets organized into four meta-tasks: classification, VQA, retrieval, and visual grounding. MMEB provides 20 in-distribution (IND) training datasets. Evaluation is conducted on a comprehensive test set of 36 datasets, including 20 IND and 16 out-of-distribution (OOD) datasets.  This diverse composition enables a rigorous assessment of both robustness and generalization across multiple tasks and domains.

\paragraph{Metrics} For evaluation, we report $Precision@1$ for each dataset following the existing work ~\cite{jiang2024vlm2vec}, which measures the proportion of top-ranked candidates that are positive samples. 
% MMEB is designed to facilitate the training and evaluation of instruction-following multimodal embedding models, with a strong emphasis on cross-task and cross-domain generalization.

\paragraph{Baselines} Following VLM2Vec ~\cite{jiang2024vlm2vec}, we experiment with different backbone models, i.e., Qwen2-VL-2B~\cite{wang2024qwen2}, LLaVA-NeXT~\cite{liu2024llavanext} and Phi3.5-V-4B~\cite{abdin2024phi}. Additionally, we benchmark our results against recent advanced models: CLIP~\cite{radford2021clip}, BLIP2~\cite{li2023blip}, UniIR~\cite{wei2024uniir}, MagicLens~\cite{zhang2024magiclens}, VLM2Vec~\cite{jiang2024vlm2vec}, UniME~\cite{gu2025breaking}, MegaPairs~\cite{zhou2024megapairs}, LLaVE~\cite{lan2025llave}. 

% More implementation settings can be found in the Appendix.

\subsection{Main Results}
From Table \ref{tab:result}, the \SDE framework consistently outperforms baselines across the MMEB benchmark. The best-performing variant, \SDE(LLaVA-1.6-HR), utilizes the LLaVA-1.6 backbone with high-resolution image inputs and achieves an average $Precision@1$ of $65.6$ across all 36 datasets, surpassing all compared methods. Notably, \SDE(LLaVA-1.6-HR) demonstrates robust performance across meta-task categories, with $Precision@1$ scores of $61.6$ for classification, $54.7$ for VQA, $69.0$ for retrieval, and $92.5$ for grounding, indicating strong capability in handling diverse multimodal tasks.

\begin{table*}[t]
\centering
\caption{
Performance comparison on the MMEB benchmark.
The FF and SF subscripts under CLIP or BLIP represent feature-level fusion and score-level fusion, respectively. LR suffixes signify training and inference on low-resolution ($336\times336$) images, while HR suffixes similarly denote both processes on high-resolution ($1344\times1344$) images. Results report the average $Precision@1$ across datasets, with top performances highlighted in bold and second-best scores underlined. The red numbers in parentheses indicate the relative improvement, computed as $(\text{SDE} - \text{VLM2Vec}) / \text{VLM2Vec}$.
}
\label{tab:result}
\resizebox{0.76\textwidth}{!}{%
\small
\tabcolsep=3pt
\begin{tabular}{lccccccc}
\toprule
\multirow{2}{*}{Model} & \multicolumn{4}{c}{Per Meta-Task Score} & \multicolumn{3}{c}{Avg Score} \\
\cmidrule(lr){2-5}
& Classification & VQA & Retrieval & Grounding & IND & OOD & Overall \\
\# Datasets & 10 & 10 & 12 & 4 & 20 & 16 & 36 \\
\midrule
\multicolumn{8}{c}{\textit{\textbf{Zero-shot on MMEB}}} \\
\hline
CLIP~\cite{radford2021clip} & 42.8 & 9.1 & 53.0 & 51.8 & 37.1 & 38.7 & 37.8 \\
BLIP2~\cite{li2023blip} & 27.0 & 4.2 & 33.9 & 47.0 & 25.3 & 25.1 & 25.2 \\
UniIR(BLIP$_{FF}$)~\cite{wei2024uniir} & 42.1 & 15.0 & 60.1 & 62.2 & 44.7 & 40.4 & 42.8 \\
UniIR(CLIP$_{SF}$)~\cite{wei2024uniir} & 44.3 & 16.2 & 61.8 & 65.3 & 47.1 & 41.7 & 44.7 \\
Magiclens~\cite{zhang2024magiclens} & 38.8 & 8.3 & 35.4 & 26.0 & 31.0 & 23.7 & 27.8 \\
\midrule
\multicolumn{8}{c}{\textit{\textbf{Fine-tuning on MMEB}}} \\
\hline
CLIP-FFT & 55.2 & 19.7 & 53.2 & 62.2 & 47.6 & 42.8 & 45.4 \\
OpenCLIP-FFT & 56.0 & 21.9 & 55.4 & 64.1 & 50.5 & 43.1 & 47.2 \\
UniME(Phi3.5-V)~\cite{gu2025breaking} & 54.6 & 55.9 & 64.5 & 81.8 & 68.2 & 52.7 & 61.3 \\
UniME(LLaVA-1.6)~\cite{gu2025breaking} & 60.6 & 52.9 & 67.9 & 85.1 & 68.4 & 57.9 & 63.6 \\
MegaPairs(LLaVA-1.6)~\cite{zhou2024megapairs}& 56.0& \underline{57.4} & \textbf{69.9}& 83.6& 68.0& \textbf{59.1} & 64.1\\
\midrule
\multicolumn{8}{c}{\textit{\textbf{SDE}} (\textit{Ours})} \\
\hline
VLM2Vec(Phi3.5-V)~\cite{jiang2024vlm2vec} & 54.8 & 54.9 & 62.3 & 79.5 & 66.5 & 52.0 & 60.1 \\
\rowcolor{blue!10}
SDE(Phi3.5-V) & 54.6 & \textbf{60.3} & 63.4 & 78.4 & \underline{69.8} & 51.6 & \textbf{61.7 \textcolor{red}{(+2.7)}}\\
VLM2Vec(Qwen2-VL-2B)~\cite{jiang2024vlm2vec} & 59.0 & 49.4 & 65.4 & 73.4 & 66.0 & 52.6 & 60.1 \\
\rowcolor{blue!10}
SDE(Qwen2-VL-2B) & 60.0 & 51.0 & 66.4 & 72.5 & 66.9 & 52.7 & \textbf{61.1 \textcolor{red}{(+1.5)}} \\
VLM2Vec(LLaVA-1.6-LR)~\cite{jiang2024vlm2vec} & 54.7 & 50.3 & 56.2 & 64.0 & 61.0 & 47.5 & 55.0 \\
\rowcolor{blue!10}
SDE(LLaVA-1.6-LR) & 57.2 & 43.6 & 63.6 & \underline{87.7} & 66.3 & 49.8 & \textbf{59.0 \textcolor{red}{(+7.3)}}  \\
VLM2Vec(LLaVA-1.6-HR)~\cite{jiang2024vlm2vec} & \underline{61.2} & 49.9 & 67.4 & 86.1 & 67.5 & 57.1 & 62.9 \\
\rowcolor{blue!10}
SDE(LLaVA-1.6-HR) &  \textbf{61.6} & 54.7  & \underline{69.0} & \textbf{92.5} &  \textbf{71.6} &  \underline{58.1} &  \textbf{65.6 \textcolor{red}{(+4.3)}} \\
\bottomrule
\end{tabular}
}
\end{table*}

In addition, compared to the best baseline model with fine-tuning, MegaPairs (LLaVA-1.6), our model shows a $1.5\%$ improvement. Furthermore, our method achieves the highest IND and OOD scores among all models, with $71.6$ and $58.1$, respectively, highlighting its generalization ability to unseen tasks.

Other \SDE variants, such as \SDE(Phi3.5-V) and \SDE(Qwen2-VL-2B), also demonstrate notable improvements over their respective VLM2Vec baselines, with gains of $+2.7$ and $+1.5$ points in overall average $Precision@1$. This consistent performance boost across different backbones and resolutions suggests that our approach is broadly effective.

\subsection{Generalization Analysis}
\begin{figure*}[ht]
\centering
 \includegraphics[width=1.0\textwidth]{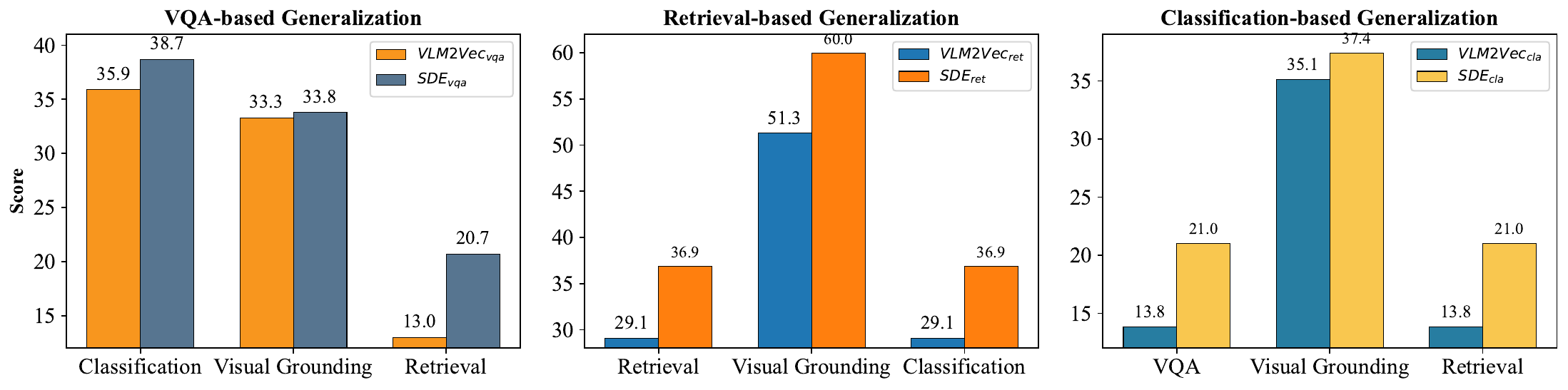}
\caption{Cross-task generalization performance comparison between VLM2Vec and \SDE. Each subplot shows the performance when trained on one meta-task, e.g., VQA, classification, or retrieval, and evaluated on other unseen tasks. Notably, \SDE demonstrates superior generalization capabilities across all scenarios. Both models employ Qwen2-VL-2B as the backbone.}
\label{fig:task_generalization}
\end{figure*}

We demonstrate in Table~\ref{tab:result} that \SDE, after being trained on a diverse set of IND tasks, can effectively generalize to OOD datasets.
A natural question is whether focusing training on a single meta-task can further improve the generalization of the model to other task types. To investigate this, we train three separate models using the \texttt{SDE} framework, each specialized for one meta-task: classification, VQA, and retrieval. Following the work \cite{jiang2024vlm2vec}, visual grounding is not included due to the limited number of training datasets. Specifically, the models are trained as follows: $\texttt{SDE}_{cla}$ is trained on $5$ classification tasks; $\texttt{SDE}_{vqa}$ is trained on $6$ VQA tasks; and $\texttt{SDE}_{ret}$ is trained on $8$ retrieval tasks. We evaluate their generalization performance on the remaining cross-task datasets.  Fig. \ref{fig:task_generalization} presents the generalization performance of each model when evaluated on unseen meta-tasks.

The results reveal that \SDE consistently outperforms the strong competitor VLM2Vec in all cross-task transfer scenarios, with substantial improvements. For example, $\texttt{SDE}_{cla}$, trained only on the classification task, improves VQA performance by 52\%, achieving a score of 21.0 compared with 13.8 for VLM2Vec, while $\texttt{SDE}_{vqa}$ enhances retrieval performance by 59\%, reaching 20.7 compared with 13.0. These results validate the robustness and generalization of the spectral disentanglement and enhancement mechanism, which enables \SDE to capture richer and more generalizable multimodal semantics.

Moreover, $\texttt{SDE}_{ret}$, trained on the retrieval task, exhibits particularly strong performance when transferred to both classification and visual grounding tasks. Specifically, it achieves a 27\% improvement on classification, reaching a score of 36.9 compared with 29.1 for VLM2Vec, and a 17\% improvement on grounding, achieving 60.0 compared with 51.3. This substantial margin highlights the effectiveness of the spectral disentanglement and enhancement mechanism in extracting transferable and robust features from diverse retrieval scenarios.

\subsection{Ablation Study}
\begin{table*}[ht]
\centering
\caption{The ablation study of the \SDE framework on the MMEB benchmark. We compare the performance of the full \SDE model with ablated variants that selectively activate different spectral components, i.e., strong, weak, and noise signals, while keeping the dual domain contrastive loss. Reported scores are the average $precision@1$ over the corresponding datasets. Here, \SDE and its variants utilize the Qwen2-VL-2B as the backbone architecture.}
\label{tab:meta_task_results}
\resizebox{0.76\textwidth}{!}{%
\small
\begin{tabular}{lccccccc}
\toprule
\multirow{2}{*}{Model} & \multicolumn{4}{c}{Per Meta-Task Score} & \multicolumn{3}{c}{Avg Score} \\
\cmidrule(lr){2-5} \cmidrule(lr){6-8}
& Classification & VQA & Retrieval & Grounding & IND & OOD & Overall \\
\# Datasets & 10 & 10 & 12 & 4 & 20 & 16 & 36 \\
\midrule

\SDE (Strong Only)              & 59.8  & 50.9  & 66.3   & \textbf{72.7}   & 67.0   & \textbf{53.3}   & 61.0 \\
\SDE (Weak Only)                & 59.0   & \textbf{51.4}  & 65.9   & 71.0   & 66.8   & 52.6   & 60.5 \\
\SDE (Noise Only)               & 59.8  &  51.1   & 66.0   & 72.0   & 67.0   & 52.7   & 60.8 \\
\textbf{\SDE (Ours)}           & \textbf{60.0}  & 51.0   & 66.4   & 72.5   & \textbf{67.1}   & 52.7   & \textbf{61.1} \\
\bottomrule
\end{tabular}
}
\end{table*}

To systematically assess the contribution of different spectral components in our framework,  we conduct an ablation study investigating the effect of each component—namely, the strong, weak, and noise signals.
%we conduct two complementary ablation studies. The first investigates the effect of each spectral component, while the second evaluates the role of different loss objectives.

As shown in Table~\ref{tab:meta_task_results}, we compare \SDE variants that selectively activate different spectral components. Each ablated variant performs competitively in specific scenarios, with strong signals excelling in visual grounding and weak signals performing best on VQA. However, none matches the full model’s consistent performance across all tasks. The complete \SDE framework achieves the highest overall accuracy and IND performance, demonstrating that integrating all spectral components is essential for optimal generalization. Notably, the strong-signal variant attains the best OOD performance, indicating that dominant features provide crucial robustness against distribution shifts, while the weak and noise components contribute to task-specific refinements.

\subsection{Qualitative Study}
\begin{figure*}[ht]
  \centering
   \includegraphics[width=0.95\linewidth]{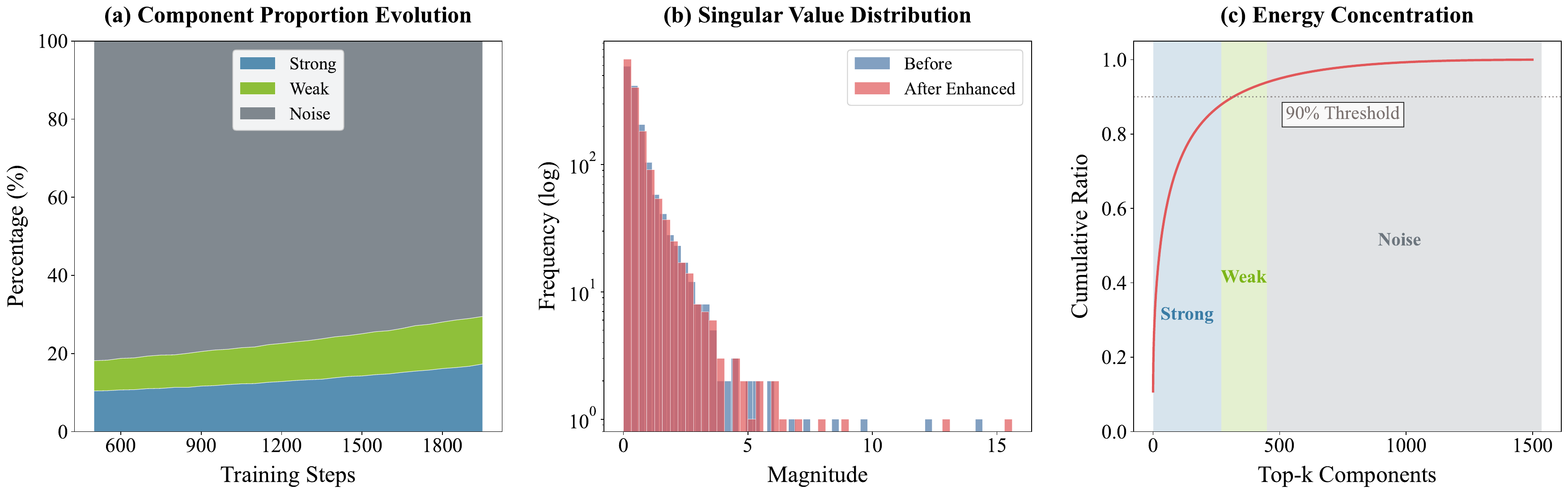}
    \caption{Qualitative examples of spectral disentanglement and enhancement. (a) Evolution of component proportions shows increasing dominance of strong signals. (b) Singular value distributions before and after enhancement, demonstrating selective amplification of meaningful features. (c) Cumulative energy distribution highlighting the concentration of semantic information in strong components.}
  \label{fig:triple}
\end{figure*}

% Fig.~\ref{fig:triple} provides compelling visual evidence of our spectral disentanglement and enhancement mechanism. In Fig.~\ref{fig:triple} (a), we observe that strong signals exhibit the most significant growth (+66.25\%), increasing from 10.42\% to 17.32\% of total components, weak signals show robust expansion (+57.14\%), rising from 7.75\% to 12.17\%, and noise components are effectively suppressed (-13.84\%), decreasing from 81.84\% to 70.51\%.
% Fig.~\ref{fig:triple} (b) demonstrates that spectral enhancement shifts the singular value distribution, the noise components (small singular values)  decrease, and mid-range singular values remain largely unchanged, and the dominant signals are amplified.
% Fig.~\ref{fig:triple} (c) shows that strong components dominate 88.7\% of the total energy, weak components contribute 4.0\% of the total energy, and noise accounts for only 7.3\% of the energy despite comprising 70.51\% of the total components.

Fig.~\ref{fig:triple} provides compelling visual evidence of our spectral disentanglement and enhancement mechanism. In Fig.~\ref{fig:triple} (a), strong signals exhibit the most significant growth, increasing by 66.25\% from 10.42\% to 17.32\% of total components. Weak signals also expand substantially, rising by 57.14\% from 7.75\% to 12.17\%, while noise components are effectively suppressed, decreasing by 13.84\% from 81.84\% to 70.51\%. Fig.~\ref{fig:triple} (b) illustrates that spectral enhancement shifts the singular value distribution: the proportion of noise components with small singular values decreases, mid-range singular values remain largely stable, and dominant signals are amplified. Fig.~\ref{fig:triple} (c) shows that strong components account for 88.7\% of the total energy, weak components contribute 4.0\%, and noise comprises only 7.3\% despite representing 70.51\% of the components.

These quantitative visualizations demonstrate that \SDE effectively reweights the spectral composition of learned representations by dynamically amplifying semantically meaningful signals while suppressing noise components. The strategy of spectral decomposition and enhancement enables \SDE to achieve optimal information compression - distilling representations into their most semantical components while maintaining expressiveness through preserved mid-range features, ultimately yielding compact yet highly discriminative embeddings.

%Strong 数量变化百分比: 66.25%
%Strong 初始占比: 10.42%
%Strong 最终占比: 17.32%
%Weak 数量变化百分比: 57.14%
%Weak 初始占比: 7.75%
%Weak 最终占比: 12.17%
%Noise 数量变化百分比: -13.84%
%Noise 初始占比: 81.84%
%Noise 最终占比: 70.51%
%\subsection{Training Parameters Analysis}
%We further investigate the effect of varying batch size on model performance and stability. Models are trained and evaluated with different batch sizes and compared with baselines from SOTA. This analysis reveals the interaction between batch-level spectral properties and the quality of representation.

%\begin{table*}[h]
%\centering
%\caption{Mean accuracy comparison of VLM2VEC and SDE under different batch sizes. Relative Improvement (\%) denotes the percentage increase in accuracy achieved by SDE over VLM2VEC.}
%\label{tab:bs_result}
%\resizebox{0.8\textwidth}{!}{%
%\begin{tabular}{lccccc}
%\toprule
%Model      & $batch\_size=128$ & $batch\_size=256$ & $batch\_size=512$ & $batch\_size=1024$ & $batch\_size=2048$ \\
%\midrule
%VLM2VEC    & 46.9  & 48.1  & 50.1  & 52.1   & 53.7   \\
%SDE        & \textbf{51.0}  & \textbf{52.4}  & \textbf{54.2} & \textbf{54.9}   & \textbf{54.7}   \\
%\midrule
%\rowcolor{blue!10}
%\textit{Relative Improvement (\%)} & \textbf{8.7}  & \textbf{8.7}  & \textbf{8.2}  & \textbf{5.4}   & %\textbf{1.9}   \\
%\bottomrule
%\end{tabular}
%}
%\end{table*}

\section{Conclusion}
\label{sec_conclu}
In this work, we present \texttt{SDE}, a novel framework that advances multimodal representation learning by explicitly modeling and optimizing the spectral properties of learned embeddings. Existing methods mainly treat the feature dimensions uniformly and ignore the spectral imbalance issue. \SDE provides an approach to disentangle and enhance representations in the spectral domain. Our approach introduces three key contributions: an adaptive spectral decomposition module that identifies and separates semantically meaningful components from noise, improving feature interpretability; a theoretically-motivated enhancement strategy that differentially amplifies informative spectral components, ensuring robustness and discriminative power; and a dual-domain learning objective that simultaneously optimizes feature-space discrimination and spectral-space consistency, leading to more transferable representations. Extensive experiments demonstrate that \SDE-learned representations achieve superior generalization across diverse multimodal tasks, outperforming existing methods on large-scale benchmarks. Our framework not only advances the theoretical understanding of spectral properties in representation learning but also provides a scalable and extensible solution for a broad range of VLMs. 

Future work could explore extending \SDE to dynamic or sequential multimodal settings, where the spectral properties of representations may evolve over time. This calls for the design of adaptive mechanisms capable of tracking and adjusting spectral decomposition in real time, thereby enhancing robustness in processing temporal data such as video or audio streams. Such an extension could further bridge the gap between static and dynamic multimodal learning, unlocking applications in areas like video understanding.

%In this work, we present Spectral Disentanglement and Enhancement (\texttt{SDE}), a novel dual-domain contrastive framework for robust and generalizable multimodal representation learning. By explicitly modeling the spectral structure of feature representations, SDE overcomes the limitations of conventional contrastive methods that treat all feature dimensions uniformly. Our approach introduces three key innovations: (1) \textit{Spectral Decomposition}, which dynamically partitions feature embeddings into strong, weak, and noise subspaces via real-time singular value decomposition; (2) \textit{Spectral Enhancement}, a curriculum-based strategy that selectively amplifies informative signals and suppresses noise, with theoretical guarantees on training stability; and (3) \textit{Dual-domain Contrastive Learning}, which jointly optimizes feature-space and spectral-space objectives to promote disentangled and diverse representations. Extensive experiments on the Massive Multimodal Embedding Benchmark (MMEB) demonstrate that \texttt{SDE} consistently outperforms state-of-the-art baselines, achieving superior robustness and improved generalization across a range of tasks and domains. Ablation studies further validate the contributions of each component, highlighting the importance of spectral-aware optimization and dual-domain alignment.


\section{Acknowledgment}
This work was partly supported by the Beijing Natural Science Foundation No. 4254079.

\bibliographystyle{ACM-Reference-Format}
\balance
\bibliography{mybib}

%\clearpage
%\pagebreak
\appendix

\section{Implementation Setting}
\begin{figure}[H]
\centering
\begin{subfigure}{0.235\textwidth}
\centering
\includegraphics[width=\linewidth]{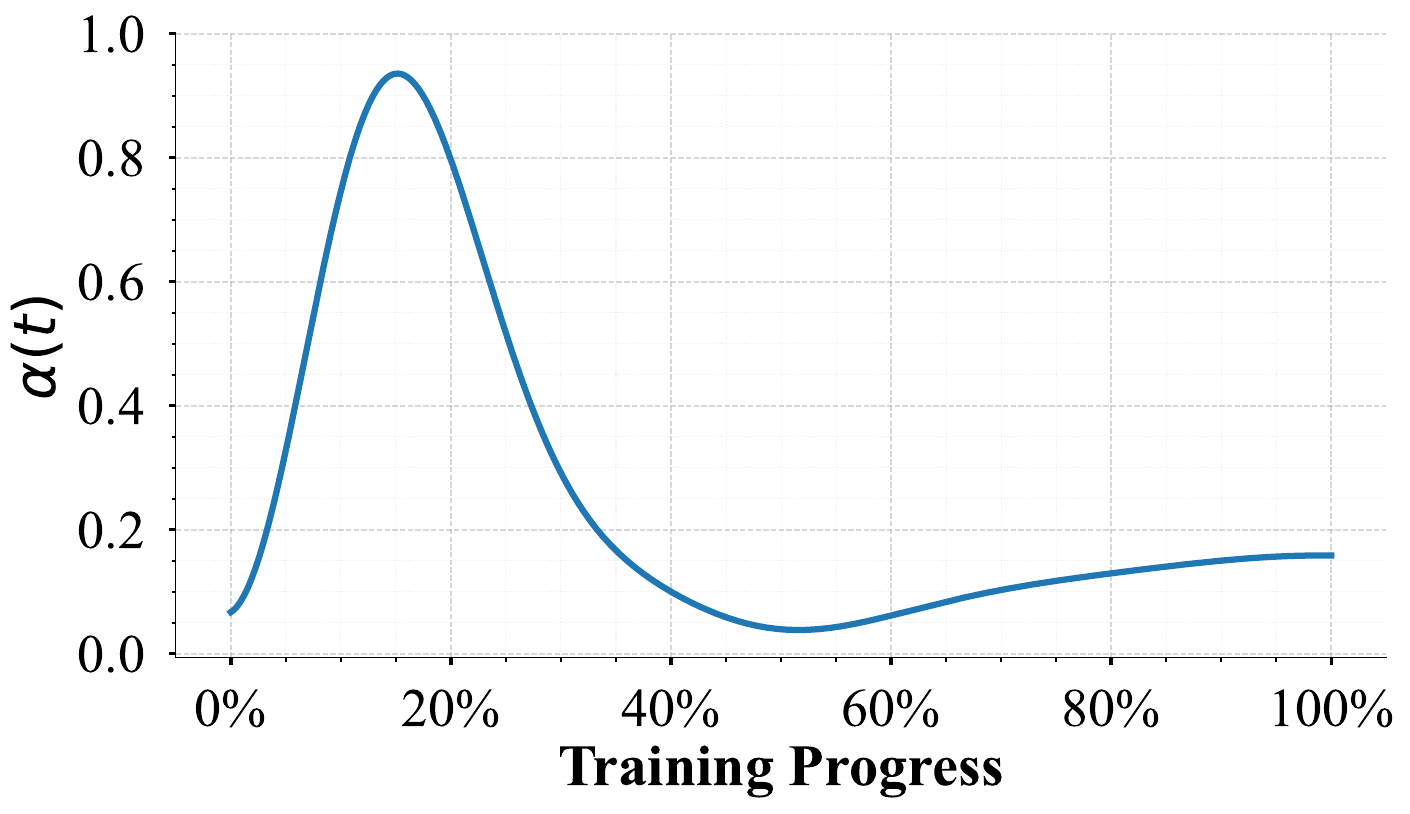}
\caption{Curriculum factor $\alpha(t)$}
\label{fig:alpha_t}
\end{subfigure}
%\hfill
\begin{subfigure}{0.235\textwidth}
\centering
\includegraphics[width=\linewidth]{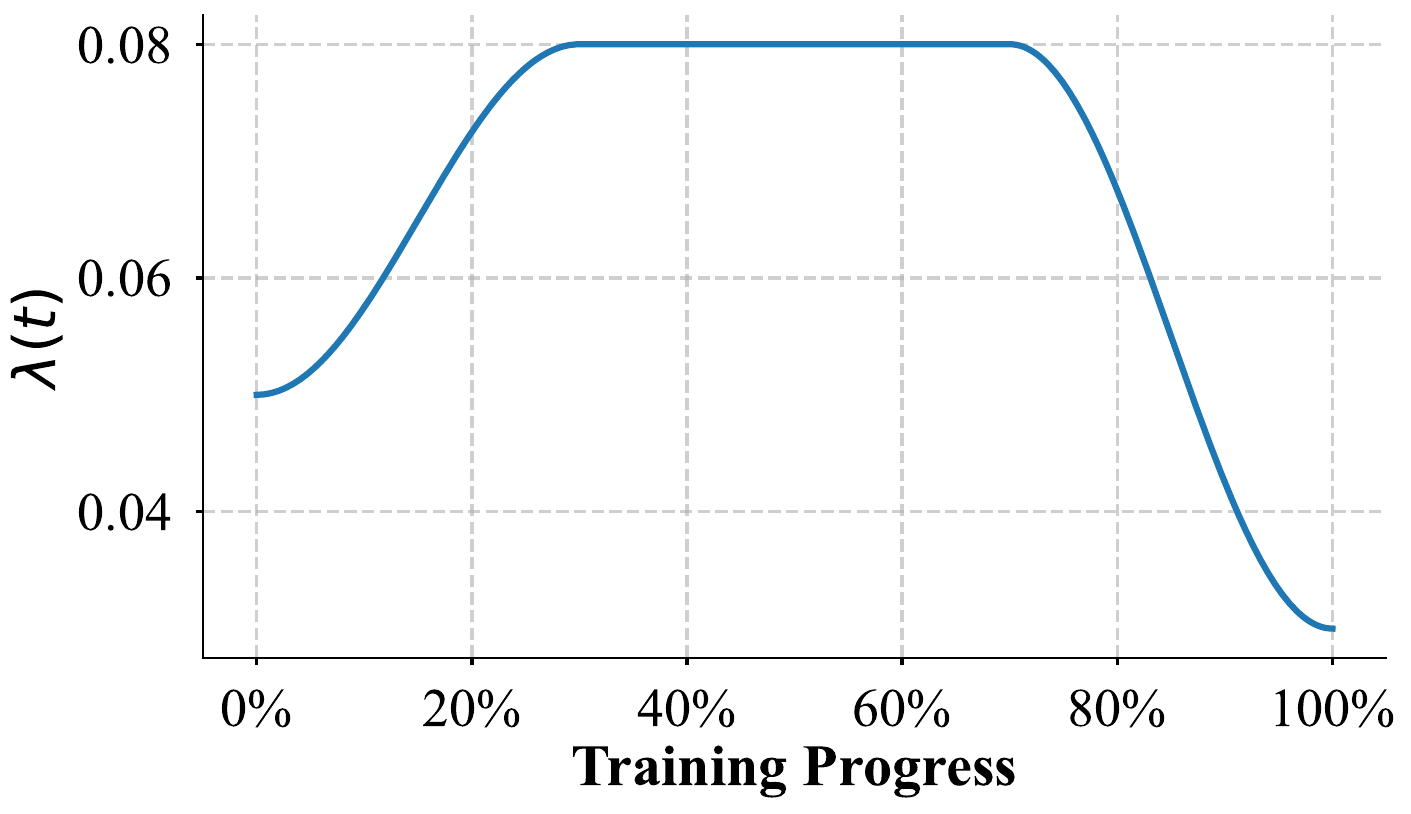}
\caption{Dynamic weight $\lambda(t)$}
\label{fig:lambda_t}
\end{subfigure}
\caption{Hyperparameter scheduling patterns during training: (a) illustrates the decay of the curriculum factor $\alpha(t)$, while (b) shows the dynamically scheduled weighting coefficient $\lambda(t)$.}
\label{fig:alpha_lambda}
\end{figure}

Our implementation builds on the Qwen2-VL-2B architecture~\cite{wang2024qwen2}, following the methodological framework of VLM2Vec~\cite{jiang2024vlm2vec}. For parameter-efficient fine-tuning, we integrate LoRA adapters~\cite{hu2022lora} with a rank of $8$ across all trainable layers. The detailed training configuration is as follows.
\begin{itemize}[leftmargin=2em]
    \item \textbf{Hardware}. Experiments are conducted on $8\times$ H800 GPUs ($80$GB VRAM per device).
    \item \textbf{Data Processing}. High-resolution images are tokenized into sequences of up to $4,096$ tokens, with a maximum of $100,000$ samples per dataset to ensure balanced resource allocation.
    \item \textbf{Optimization}. Optimization is performed using a linear learning rate scheduler, with an initial rate of $2\times 10^{-5}$ over $2,000$ training steps, including $200$ warmup iterations.
    \item \textbf{Batch Handling}. Each GPU processes a local batch of 256 samples. To mitigate memory constraints, we leverage GradCache~\cite{gao2021GradCache} with chunk sizes of $4$ for both query and candidate embeddings, enabling efficient gradient computation across distributed units.
    \item \textbf{Hyperparameter setting}. The temperature $\tau$ in Eq.~\ref{eq:infonce} is consistently set to $0.02$ across different VLM scales. The curriculum factor $\alpha(t)$ in Eq.~\ref{eq:delta} is defined as a piecewise function of training step $t$, where $T$ is the total steps and $p=t/T$ indicates the training progress:
    $$
     \alpha(t) = 
    \begin{cases}
    (0.8 - 0.15\beta)\left[1 - \cos(6\pi t/T)\right], & p < 0.15 \\
    (0.4 - 0.08\beta)\left[1 + \cos(3\pi(t/T - 0.15))\right], & 0.15 \leq p < 0.5 \\
    (0.1 - 0.02\beta)\left[1 - \cos(2\pi(t/T - 0.5))\right], & p \geq 0.5
    \end{cases}
    $$
    where $\beta = \log(\text{batch\_size}/256 + 1)/\log(8)$ is the batch scaling factor. This formulation implements a curriculum learning strategy with strong augmentation in early stages ($p<0.15$), moderate augmentation in middle stages ($0.15\leq p<0.5$), and weak augmentation in final stages ($p\geq 0.5$).

    The dynamically scheduled weight $\lambda(t)$ in Eq.~\ref{eq:total_loss} follows a piecewise cosine schedule defined over normalized training progress. It is given by:
    $$
\lambda(t) =
\begin{cases}
  0.05 + 0.015 \left(1 - \cos\left(\dfrac{\pi t}{0.3}\right)\right), & 0 \leq p < 0.3 \\
        0.08, & 0.3 \leq p < 0.7 \\
        0.08 - 0.025 \left(1 - \cos\left(\dfrac{\pi (t-0.7)}{0.3}\right)\right). & 0.7 \leq p \leq 1
\end{cases}
$$
Such a design stabilizes training by emphasizing the spectral loss in the middle phase while reducing its influence toward the end. The evolution of both $\alpha(t)$ and $\lambda(t)$ throughout training is illustrated in Figure~\ref{fig:alpha_lambda}.
\end{itemize}

\section{Complexity Discussion}
To address computational efficiency concerns, our SVD-based decomposition is performed on a globally gathered batch, ensuring all GPUs process identical unified data and avoid fragmentation. Empirical benchmarks demonstrate that SDE introduces minimal overhead compared to VLM2VEC:
\begin{itemize}[leftmargin=2em]
\item \textbf{Time Overhead}: Only $+2.87\%$ per-step time (195.26s → 200.87s)
\item \textbf{Throughput}: Merely $-3.05\%$ decrease (1.31 → 1.27 samples/s)
\item \textbf{Memory}: Nearly zero additional usage
\end{itemize}
The overhead remains below $3\%$ across batch sizes from 64 to 1024, confirming that \SDE scales efficiently without introducing computational or memory bottlenecks. This validates the practical viability of our approach in large-scale distributed training scenarios.

\section{Ablation Study on Loss Objectives}
We empirically evaluate the effectiveness of our dual-domain contrastive learning loss by comparing several \SDE variants trained with different loss objectives. To measure resilience to input perturbations, we randomly perturb 5\% of each test dataset using orthogonal transformations (Eq.~\ref{eq:orth_trans}).

\begin{table}[H]
\centering
\caption{The ablation study of the \SDE framework and its variants with different loss objectives on the MMEB benchmark. For example, \SDE-feat means that it only employs instance-level contrastive loss in the feature space. Here, \SDE and its variant utilize Qwen2-VL-2B as the backbone with a small batch size of 16 for efficiency. }
\label{tab:loss_results}
\resizebox{0.48\textwidth}{!}{%
\begin{tabular}{lcccccc}
\toprule
\multirow{2}{*}{Model} & \multirow{2}{*}{Loss Objective} & \multicolumn{2}{c}{Avg Score} \\
\cmidrule(lr){3-4}
& & Overall & Perturbed overall(5\%) & & \\
\midrule
\SDE-feat   & $\mathcal{L}_{\text{feat}}$ & 45.8 & 41.9  \\
\SDE-S      & $\mathcal{L}_{\text{feat}} + \mathcal{L}_{\text{hellinger}}$ & 46.0  & 42.6   \\
\SDE-U      & $\mathcal{L}_{\text{feat}} + \mathcal{L}_{\text{subspace}}$ & 46.2 & 43.0    \\
\textbf{\SDE (Ours)} & $\mathcal{L}_{\text{feat}} + \mathcal{L}_{\text{spec}}$ & \textbf{48.8}  & \textbf{47.8} \\
\bottomrule
\end{tabular}
}
\end{table}

As shown in Table~\ref{tab:loss_results}, the full \SDE model, which incorporates both instance-level and spectral structure alignment through the composite spectral contrastive loss $\mathcal{L}_{feat} + \mathcal{L}_{spec}$, achieves the highest $Precision@1$ on clean data at 48.8 and maintains superior performance under input perturbation, reaching 47.8, with the smallest performance degradation of 2.0\%. The feature-level alignment term $\mathcal{L}_{feat}$ alone shows significant vulnerability to perturbations, exhibiting an 8.5\% drop. Spectral components provide increasing robustness, as demonstrated by the \textit{Hellinger} loss with a 7.4\% drop and the \textit{Subspace} loss with a 6.9\% drop. These results demonstrate that feature-level discrimination and spectral-level consistency offer complementary advantages, and their combination yields both high accuracy and enhanced robustness.

%\section{Visual Evidence}
%\begin{figure*}[htbp] % 位置参数：h=此处, t=顶部, b=底部, p=单独一页
%  \centering 
%  \includegraphics[width=0.95\textwidth]{feat.pdf} 
%  \caption{xx} 
%  \label{fig:feat} 
%\end{figure*}

\end{document}